\documentclass[letterpaper, 10 pt, conference]{ieeeconf}  

\IEEEoverridecommandlockouts                              
\usepackage{cite}
\usepackage{graphicx} 
\usepackage{comment}
\usepackage{amsmath, amssymb, algorithm, algorithmic}

\title{\LARGE \bf Platform-Agnostic Reinforcement Learning Framework for Safe Exploration of Cluttered Environments with Graph Attention}

\author{Gabriele Calzolari, Vidya Sumathy, Christoforos Kanellakis, George Nikolakopoulos  
\thanks{This work was partially supported by the Wallenberg AI, Autonomous Systems and Software Program (WASP) funded by the Knut and Alice Wallenberg Foundation, and by the European Union's Horizon Europe Research and Innovation Program, under the Grant Agreement No. 101119774 SPEAR. This research was conducted using the resources of High Performance Computing Center North (HPC2N). Additionaly, the RL-training were enabled by resources provided by the National Academic Infrastructure for Supercomputing in Sweden (NAISS), partially funded by the Swedish Research Council through grant agreement no. 2022-06725. The authors are within the Robotics and AI Group, Department of Computer Science, Electrical and Space Engineering, Luleå University of Technology, Sweden. Corresponding author's e-mail: gabcal@ltu.se}}

\begin{document}

\maketitle
\thispagestyle{empty}
\pagestyle{empty}

\begin{abstract}

Autonomous exploration of obstacle-rich spaces requires strategies that ensure efficiency while guaranteeing safety against collisions with obstacles. This paper investigates a novel platform-agnostic reinforcement learning framework that integrates a graph neural network-based policy for next-waypoint selection, with a safety filter ensuring safe mobility. Specifically, the neural network is trained using reinforcement learning through the Proximal Policy Optimization (PPO) algorithm to maximize exploration efficiency while minimizing safety filter interventions. Henceforth, when the policy proposes an infeasible action, the safety filter overrides it with the closest feasible alternative, ensuring consistent system behavior. In addition, this paper introduces a reward function shaped by a potential field that accounts for both the agent’s proximity to unexplored regions and the expected information gain from reaching them. The proposed framework combines the adaptability of reinforcement learning-based exploration policies with the reliability provided by explicit safety mechanisms. This feature plays a key role in enabling the deployment of learning-based policies on robotic platforms operating in real-world environments. Extensive evaluations in both simulations and experiments performed in a lab environment demonstrate that the approach achieves efficient and safe exploration in cluttered spaces.
\end{abstract}

\section{Introduction}

Efficient and safe autonomous exploration in obstacle-rich environments, such as forests and mines, remains a critical challenge for robotic systems. These issues encompass optimal path planning, uniform coverage, and ensuring safety during decision making. Traditional exploration strategies, such as random exploration and frontier-based approaches, are often inadequate in real-world environments because they lack adaptability to evolving scenarios, suboptimal path planning, and take myopic decisions. On the other hand, Reinforcement Learning (RL)-based policies enable adaptive decision-making by optimizing long-term rewards and dynamically responding to environmental changes. Among these learning-based methods, Graph Neural Network (GNN)-based architectures have gained increasing attention due to their ability to represent spatial relationships within an environment using graph structures, allowing policies to extract and leverage meaningful features from critical locations in the scene when making navigation decisions. Unlike classical strategies, learned policies provide limited insight into the decision-making process and offer no explicit guarantees that they have acquired the correct behavior beyond statistical evaluation in testing environments. This issue is especially critical in safety-sensitive applications, such as flying a drone in a thick forest, where an agent must demonstrably adhere to strict constraints to prevent hazardous behavior. To address this, recent research is focusing on integrating safety mechanisms into RL-based frameworks, ensuring compliance with safety constraints while maintaining adaptability. One promising direction involves combining deep learning methods, capable of encoding complex behaviors, with safety filters that impose hard constraints on action selection, thereby preventing unsafe decisions. This hybrid approach improves the robustness and reliability of learned policies, facilitating their deployment in real-world environments.

\begin{figure*}[h]
    \centering
    \includegraphics[width=\textwidth]{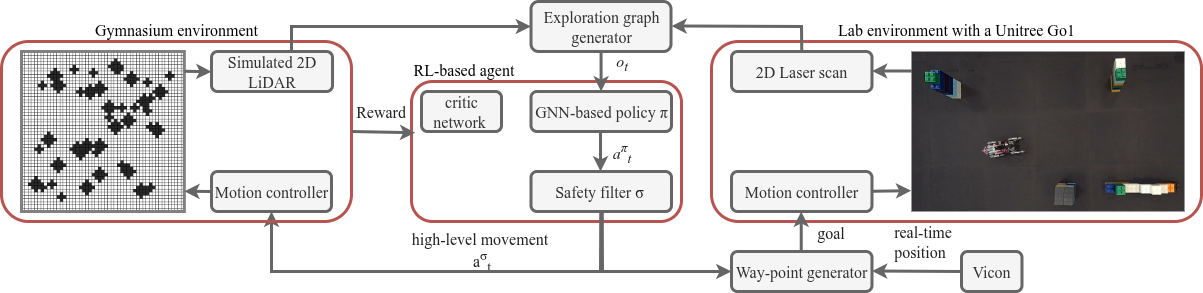} 
    \caption{Overview of the proposed safe reinforcement learning framework for exploration in cluttered environments. The hierarchical architecture integrates a GNN-based policy with a safety filter to generate the next-step waypoint for exploration. The resulting high-level movement can be applied either to a simulated environment in Gymnasium for policy training and ablation studies (left branch), or to the laboratory setting with the Unitree Go1 quadruped robot for physical experiments (right branch).}
    \label{fig:framework_overview}
\end{figure*}

\subsection{Related works}
Recently, reinforcement learning has had a profound impact on the development of autonomous robotic exploration approaches, enabling agents to safely navigate and explore cluttered environments achieving great efficiency  \cite{amin2021survey, garaffa2021reinforcement, hutsebaut2022hierarchical, singh2022reinforcement, pateria2021hierarchical,ladosz2022exploration}.
For instance, \cite{xue2022uav} investigates a Deep Reinforcement Learning-based navigation approach for unmanned aerial vehicles that integrates a stochastic value function. By formulating navigation as a partially observable Markov decision process (POMDP), the method leverages the fast recurrent stochastic value-gradient algorithm to achieve safe and reliable exploration in large, cluttered three-dimensional environments. On the other hand, \cite{cimurs2021goal} proposes a strategy for  target-oriented exploration in which waypoints are selected using a TD3-based DRL policy. This framework mitigates convergence to local optima while operating without prior maps or human intervention. Notably, \cite{cao2023ariadne} introduce ARiADNE which is an attention-based framework that allows real-time, non-myopic path planning by capturing spatial dependencies and predicting exploration gains. \cite{sun2022fully} proposes ReLMM that allows robots to learn how to navigate via reinforcement learning, using modular policies and uncertainty-driven exploration.
Notably, literature shows that a recent line of research explores the integration of Graph Neural Networks with reinforcement learning for the implementation of exploration policy \cite{liu2023graph, munikoti2023challenges, nie2023reinforcement}. Such approaches effectively exploit the underlying spatial relationships between relevant locations in the environment, thereby enhancing decision-making during exploration. Indeed, \cite{zhang2021autonomous} investigates a Spatiotemporal Neural Network on graph for efficient exploration relying on past trajectories, obstacles, and waypoints to estimate optimal targets. \cite{herrera2023learning} presents a GNN-based method for autonomous exploration which generalizes across graph topologies and ensures high performances. The work proposed by \cite{yang2022generalized} presents SGRL, which is a GNN-based deep Q-learning algorithm for autonomous driving decisions that leverages agent interactions. \cite{lu2021mgrl} proposes MGRL, a reinforcement learning-based visual navigation framework that models uncertainty through a Markov network, employs Graph Neural Networks for inference, and incorporates knowledge graphs to enhance generalizability and adaptability. To enhance the safety guarantees of reinforcement learning frameworks, Safe RL incorporates safety constraints via constrained optimization, risk-sensitive policy design, and robustness techniques, thereby promoting optimal decision-making while mitigating unsafe actions \cite{gu2024review, brunke2022safe, zhao2023state, xu2022trustworthy, guerrier2024learning}. For instance, \cite{achiam2017constrained} proposes Constrained Policy Optimization (CPO), an algorithm that enforces constraint satisfaction during policy learning while simultaneously optimizing performance, thereby enabling neural networks to address high-dimensional tasks with formal safety guarantees. \cite{yang2022cup} presents the Conservative Update Policy (CUP) algorithm, a safe reinforcement learning approach that provides theoretical safety guarantees by leveraging surrogate functions and establishing tighter performance bounds. \cite{huang2023safedreamer} investigates SafeDreamer, an approach that integrates Lagrangian methods into the Dreamer framework for safe decision-making, achieving near-zero-cost performance in vision-only and low-dimensional tasks in the Safety-Gymnasium benchmark. \cite{feng2023dense} introduces D2RL, a learning-based approach that accelerates autonomous vehicle safety testing by densifying training data around critical states, thereby expediting evaluation while preserving statistical unbiasedness. Augmented Proximal Policy Optimization (APPO) is proposed by \cite{dai2023augmented} and is a safety-constrained RL approach that improves convergence and cost control by adding a quadratic penalty to the Lagrangian function. \cite{zhang2023evaluating} proposes Unrolling Safety Layer (USL), a method combining safety optimization and projection to enforce state-wise safety constraints in model-free RL. Finally, \cite{ma2022conservative} presents the Conservative and Adaptive Penalty (CAP) framework, a model-based Safe Reinforcement Learning method that dynamically adjusts an uncertainty-aware penalty to balance reward maximization and cost minimization, thereby ensuring safety in both tabular and high-dimensional environments.

Inspired by prior studies, we investigate a novel platform-agnostic framework for safe autonomous exploration in cluttered environments. The exploration policy follows a hierarchical two-stage process: first, a Graph Neural Network, trained with reinforcement learning, selects a candidate waypoint; then, a safety filter evaluates its feasibility and, if necessary, adjusts it to ensure collision-free navigation. The most significant contributions of this work are as follows.

\begin{itemize} 
\item A platform-agnostic hierarchical framework based on reinforcement learning that combines a GNN-driven exploration policy with a novel safety filter, which intervenes by replacing motions violating safety constraints with feasible ones.

\item A novel attention-enhanced graph neural network that implements the exploration policy by extracting exploration-relevant features from a custom graph-based representation of the environment, which encodes waypoints, frontiers, and occluded regions.

\item A proposed reward function that promotes consistent map expansion using potential field variations tied to frontier information gain and distance, validated through ablation studies.

\item A validation of the proposed framework through simulations and experiments in a lab environment with a Unitree Go1 robotic platform.
\end{itemize}

\begin{figure*}[h]
    \centering
    \includegraphics[width=\linewidth]{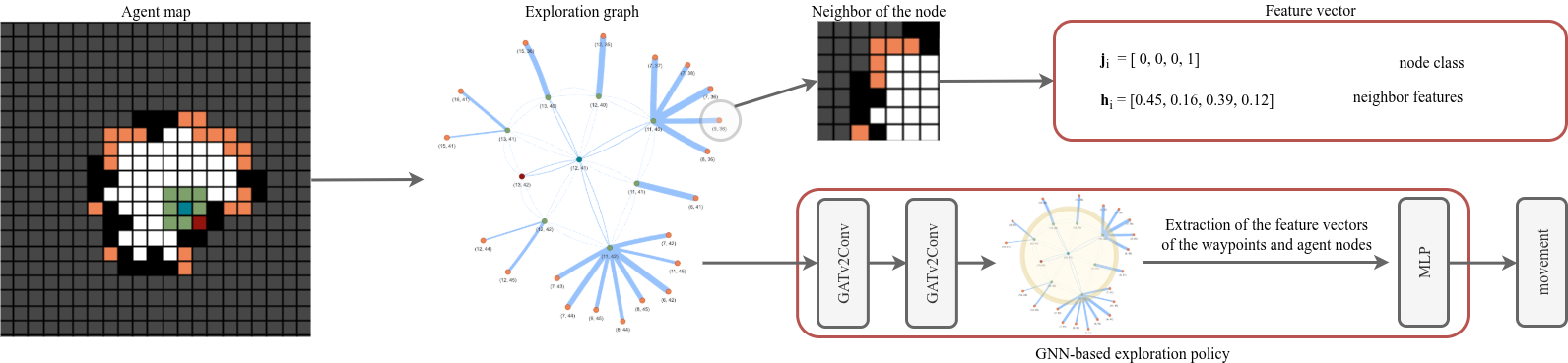} 
    \caption{Illustration of the exploration graph used as the observation $o_t$ and the GNN-based policy. The image shows a section of the agent's map, where occupied, unknown, and free cells are represented in black, gray, and white, respectively. The agent's position, feasible and infeasible next-step navigation goals, and frontiers are indicated in blue, green, red, and orange, respectively. The graph structure encodes node relationships, with edge thickness proportional to the distance between connected nodes. On the right, the local neighborhood of a frontier and its extracted feature vector are highlighted, together with the internal architecture of the exploration policy, where the main layers are emphasized.}
    \label{fig:exploration_graph}
\end{figure*}

\section{Methodology}
\label{sec:methodology}

This section presents the exploration framework formulation in \ref{sbs:agent_policy}, the design of the agent's observation space in \ref{sbs:graph_based_observation}, the GNN architecture implementing the policy and critic in \ref{sbs:gnn_architecture}, and the reward function formulation in \ref{sbs:reward_abl}.

\subsection{Platform-agnostic exploration framework}
\label{sbs:agent_policy}
The overall structure of the proposed hierarchical framework operating in obstacle-rich environments is depicted in Fig. \ref{fig:framework_overview}. Specifically, the agent is tasked to explore an arena that is represented as an occupancy grid map $\mathcal{M}_{h \times w}$, with grid resolution $r_g$, containing $n_o$ circular non-traversable regions $\mathcal{O} = \{o(c_i, r_i)\}_{i=1}^{n_o}$, with radii $r_i \sim \mathcal{U}[r_{\min}, r_{\max}]$ and centers $c_i$ randomly distributed in the environment. At each time step $t$, the exploring agent occupies a grid cell $v_t^a$ and has a partial map $\mathcal{M}^a_t$ of the reconstructed environment that contains only explored regions. Upon reaching a goal, the agent updates $\mathcal{M}^a_t$ using simulated 360-degree LiDAR measurements with range $l$. From this map, a graph-based observation $o_t$ is constructed for the policy, encoding the agent’s position, navigation goals, and the frontiers between traversable and unknown regions, as described in \ref{sbs:graph_based_observation}. 

As illustrated in Fig. \ref{fig:framework_overview}, the hierarchical exploration strategy combines an RL-trained policy $\pi_\theta(o_t)$, which selects a movement $a^\pi_t \in \mathcal{A}$ towards one of the eight neighboring cells, with a safety filter $\sigma(a^\pi_t)$ that ensures that the agent's movement is feasible. The agent’s action space comprises the movements to its adjacent eight cells in the grid and a null action corresponding to no displacement. Formally, if the agent is at position $(i,j) \in \mathcal{M}_{h \times w}$, then an action $a = (\Delta i, \Delta j) \in \mathcal{A}$ produces the transition
\[
(i,j) \;\mapsto\; (i+\Delta i,\, j+\Delta j),
\]
where the action space is defined as
\[
\mathcal{A} = \{ (\Delta i,\Delta j) \mid \Delta i,\Delta j \in \{-1,0,1\} \}.
\]
This yields $|\mathcal{A}| = 9$, accounting for radial (axial), diagonal, and null motions of the agent. In particular, the safety filter validates the policy-selected action $a^\pi_t$ considering the agent’s reconstructed map. If the action is found feasible, then it is executed unchanged. Otherwise, the filter substitutes $a^\pi_t$ with the closest feasible action $a^{\sigma}_t \in \mathcal{A}^f_t$, determined by minimizing the angular deviation from $a^\pi_t$, i.e., by selecting the feasible action that maximizes the cosine similarity with the original policy action, as in Eq. \eqref{eq:safety_filter}.

\begin{equation}
a^{\sigma}_t =
\begin{cases}
a^\pi_t, & \text{if } a^\pi_t \in \mathcal{A}^f_t \\
\displaystyle \arg\max_{a \in \mathcal{A}^f_t} \,
\frac{\langle a^\pi_t, a \rangle}{\|a^\pi_t\| \,\|a\|}, 
& \text{otherwise}
\end{cases}
\label{eq:safety_filter}
\end{equation}

where $\langle \cdot, \cdot \rangle$ denotes the scalar product, $\mathcal{A}^f_t \subseteq \mathcal{A}$ indicates the feasible action set containing all actions that keep the agent within the map boundaries, avoid collisions with non-traversable cells, and, for diagonal moves, require that at least one of the adjacent orthogonal moves is traversable to prevent corner-cutting collisions. This safety check guarantees that the action selected will not cause a collision. The high-level movement $a^\sigma_t$ is then applied to move the agent to a new location, and this process is repeated until the exploration ratio satisfies $\rho_t \geq \rho^*$ or the number of iterations reaches $n_s \geq n_s^*$, where $\rho^*$ denotes the exploration threshold and $n_s^*$ the maximum allowed number of agent--environment interactions. At each timestep, the environment provides a reward $r_t(\mathcal{M}^a_t, v_t^a, a^\pi_t, a^\sigma_t)$ that is used during training to update the policy $\pi_\theta$ and the critic network $V_\phi$ through the PPO algorithm \cite{schulman2017proximalpolicyoptimizationalgorithms}. Notably, the proposed formulation is platform-agnostic since the exploration policy operates on a graph-based representation of the environment and outputs high-level moves that do not depend on the platform-specific dynamics.

\subsection{Agent's observation space}
\label{sbs:graph_based_observation}

At time \(t\), the agent’s observation is represented as an exploration graph $o_t = (\mathcal{V}_t, \mathcal{E}_t)$ as illustrated in Fig. \ref{fig:exploration_graph} where the node set \(\mathcal{V}_t\) consists of the next-step waypoints \(\mathcal{V}^n_t\) as per Eq. \eqref{eq:vn}, and exploration frontiers \(\mathcal{V}^f_t\). Specifically, $\mathcal{V}^f_t$ is the subset of free cells in $\mathcal{M}^a_t$ that are adjacent, under 8-connectivity, to at least one unknown cell, and represent the interface between explored and unexplored regions of the environment.

\begin{equation}
\label{eq:vn}
    \mathcal{V}^n_t = \{ v^a_t + a \;\mid\; a \in \mathcal{A} \}
\end{equation}
The edge set $\mathcal{E}_t$ encodes the spatial relations among nodes, with each edge weighted by the Euclidean distance between the centroids of the corresponding cells. Specifically, it contains bidirectional edges among all the waypoint nodes $\mathcal{V}^n_t$ and unidirectional edges connecting each frontier node $f \in \mathcal{V}^f_t$ to its nearest next-step waypoint $w^* \in \mathcal{V}^n_t$, defined by Eq. \eqref{eq:closest_nav}.
\begin{equation}
\label{eq:closest_nav}
w^* = \arg \min_{w \in \mathcal{V}^n_t} \| f - w \|_2,
\end{equation}
where \(\| \cdot \|_2\) denotes the Euclidean distance. Each node \( v_i \in \mathcal{V}_t \) is associated to a feature vector \(\mathbf{x}_i = [\, \mathbf{j}_i \;\; \mathbf{h}_i \,]\), encoding both its class and local environment descriptors derived from the current agent’s map. As illustrated in Fig. \ref{fig:exploration_graph}, the first four dimensions of the node feature vector \(\mathbf{j}_i \in \{0,1\}^4\) employ one-hot encoding to indicate whether a node corresponds to the agent's position, a traversable or non-traversable next-step waypoint, or an exploration frontier. The remaining of the feature vector \(\mathbf{h}_i \in [0,1]^4\) encodes local occupancy statistics from a $(2k+1)\times(2k+1)$ neighborhood in the agent’s map \(\mathcal{M}^a_t\) of the node $v_i$. Specifically, the normalized counts of free, unknown, occupied, and frontier cells surrounding \(v_i\) are computed according to Eq. \eqref{eq:occ_stats}.

\begin{equation}
\label{eq:occ_stats}
\mathbf{h}_i(v_i) = \frac{1}{(2k+1)^2}
\begin{bmatrix}
\Gamma_{traversable}(v_i) \\
\Gamma_{unkown}(v_i) \\
\Gamma_{non-traversable}(v_i) \\
\Gamma_{frontiers}(v_i)
\end{bmatrix}^{\!\top}
\end{equation}

where $\Gamma_{\text{type}}$ denotes the number of cells of a given type within a $(2k+1)\times(2k+1)$ neighborhood centered at cell $v_i$, defined as
\begin{equation}
\label{eq:gamma}
\Gamma_{\text{type}}(v_i) \;=\; 
\sum_{x = v_{i,x}-k}^{v_{i,x}+k} 
\sum_{y = v_{i,y}-k}^{v_{i,y}+k} 
\mathbf{1}_{type}(x,y),
\end{equation}
with $v_i=(v_{i,x}, v_{i,y})$ and $\mathbf{1}_{type}(x,y)$ is the indicator function.

\subsection{Architecture of the exploration policy and critic network}
\label{sbs:gnn_architecture}

\begin{table}[t]
    \centering
    \caption{Hyperparameters of the first-stage graph neural network layers used in both the policy and critic models}
    \label{tab:gnn_architecture}
    \begin{tabular}{|c|c|c|c|c|c|}
    \hline
        GATv2Conv & Input dim & Output dim& Heads & Activation \\
        \hline
        Layer 1 & 8 & 16 & 4 & ReLU \\
        Layer 2 & 64 & 1 & 1 & - \\
        \hline
    \end{tabular}
\end{table}

Both the exploration policy $\pi_\theta$, shown in Fig. \ref{fig:exploration_graph}, and the critic network $V_\phi$ are modeled as graph neural networks that leverage attention-based message passing. They operate on the same exploration graph $o_t$ and adopt a two-stage architecture. The first stage, identical in structure but instantiated as independent networks, encodes local relational dependencies within the graph through attention mechanisms, with the policy and critic trained separately under disjoint parameterizations. The second stage differs between the two networks: for $\pi_\theta$, it maps node embeddings to action probabilities, 
whereas for $V_\phi$, it maps the graph-level representation to a scalar value estimate, both realized through fully connected layers.

\begin{itemize}

    \item \textbf{First stage.} This network consists of two stacked GATv2Conv layers \cite{brody2022attentivegraphattentionnetworks} with hyperparameters shown in Table \ref{tab:gnn_architecture}. 
    The first layer projects the 8-dimensional node features of the exploration graph $o_t$ into a 16-dimensional hidden representation $\mathbf{x}'_i$ by using four attention heads, while integrating edge attributes into the attention mechanism as per Eq.~\ref{eq:gat_update}. Subsequently, the second layer further refines these representations using a single attention head, yielding the embedding $\mathbf{x}''_i$.    
    Formally, the embedding of a graph node after the application of one attention head of a GATv2Conv layer is computed according to Eq. \eqref{eq:gat_update}
    
    \begin{equation}
    \label{eq:gat_update}
    \mathbf{x}'_i 
    = \sum_{j \in \mathcal{N}(i) \cup \{ i \}}
    \alpha_{i,j}\,\mathbf{\Theta}_{t}\,\mathbf{x}_{j},
    \end{equation}
    
    where $\mathcal{N}(i)$ denotes the neighbors of $i$, 
    $\mathbf{\Theta}_{t}$ is a learned projection matrix, and the attention coefficients $\alpha_{i,j}$ are obtained by applying Eq. \eqref{eq:gat_edge_attention}.
    
    \begin{equation}
    \label{eq:gat_edge_attention}
    \alpha_{i,j} =
    \frac{
    e^{\mathbf{a}^{\top}\epsilon_{i,j}}}
    {\sum_{k \in \mathcal{N}(i) \cup \{ i \}}
    e^{\mathbf{a}^{\top}\epsilon_{i,k}}},
    \end{equation}
    
    with
    
    \begin{equation}
        \epsilon_{p,q} = \xi\!\left(
    \mathbf{\Theta}_{s}\mathbf{x}_p
    + \mathbf{\Theta}_{t}\mathbf{x}_q
    + \mathbf{\Theta}_{e}\mathbf{e}_{p,q}
    \right)
    \end{equation}
    where $\mathbf{a}$ is the attention vector, $\mathbf{e}_{p,q}$ is the weight of the edge connecting nodes $p$ and $q$, $\xi(\cdot)$ denotes the LeakyReLU activation function, and $\mathbf{\Theta}_{s}$, $\mathbf{\Theta}_{t}$, $\mathbf{\Theta}_{e}$ are learnable transformation matrices for source, target nodes, and the projection matrix mapping edge features into the attention space, respectively.

    \item \textbf{Second stage}. In the policy network, the output is reduced to a scalar per node, and the logits of the nodes in $\mathcal{V}_n$ define the action distribution. In the critic network, the output is 16-dimensional, averaged across the nodes in $\mathcal{V}_n$, and passed through a fully connected layer to predict the critic value.

\end{itemize}

\begin{figure}[t]
    \centering
    \includegraphics[width=\linewidth]{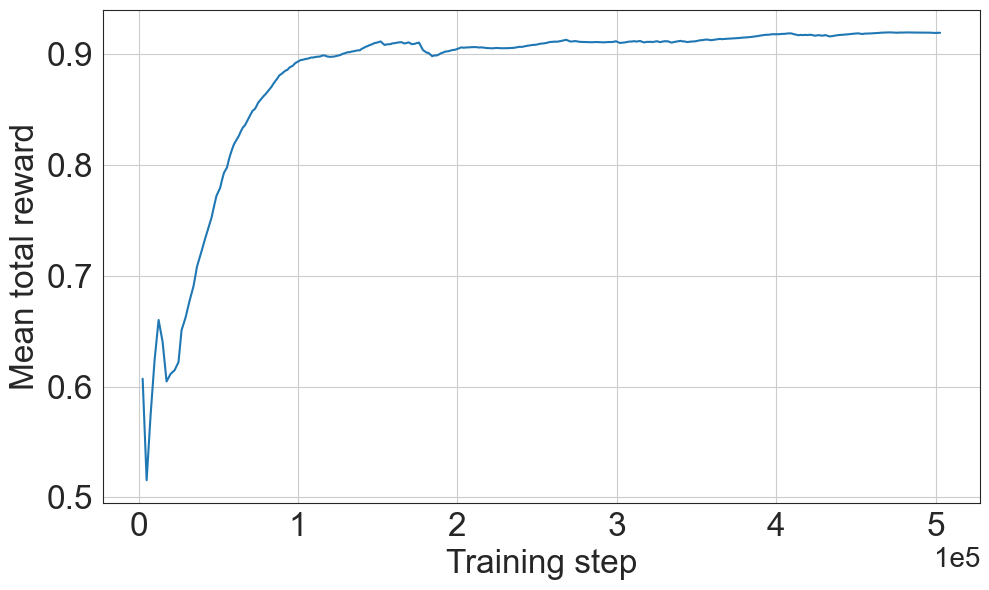}
    \caption{Training performance curve showing the mean total reward versus training steps, with rewards normalized to $[0,1]$ and smoothed using a 1000-step simple moving average.}
    \label{fig:mean_total_reward}
\end{figure} 

\begin{figure*}[t]
    \centering
    \includegraphics[width=\textwidth]{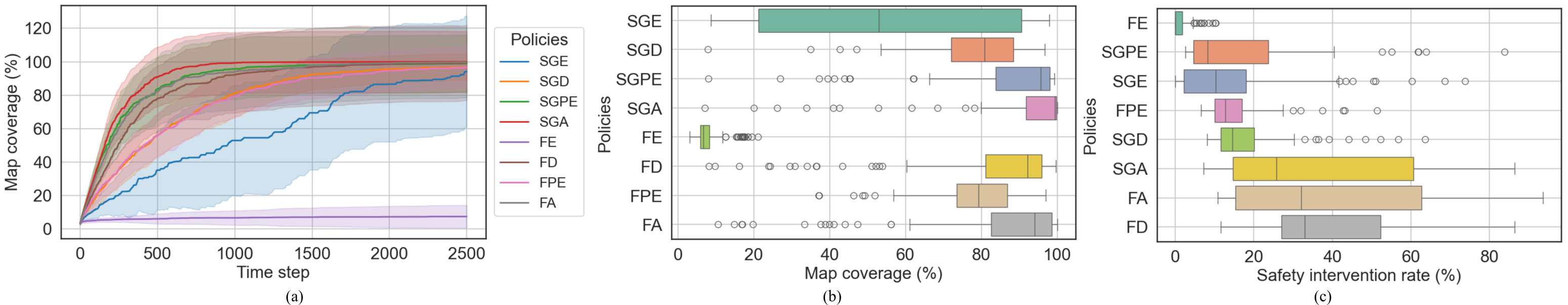}
    \caption{Data collected from the simulation of the trained policies on 100 randomly generated Gymnasium environments. From left to right: (a) Median map coverage per time step (colored lines) with variability shown as ±1 standard deviation (shaded regions) across testing environments. (b) Distribution of agents' map coverage after 1000 steps. (c) Distribution of the proportion of safety-filter interventions over agent actions across 100 test simulations per policy.}
    \label{fig:sim}
\end{figure*}

\subsection{Reward formulation and ablation study}
\label{sbs:reward_abl}

The reward function used to train the exploration policy $\pi_\theta$ is designed to encourage efficient exploration while penalizing interventions by the safety filter. We consider the base reward structure in Eq. \eqref{eq:basic_reward} to formulate the reward function.

\begin{equation} \label{eq:basic_reward}
r_t(\mathcal{M}^a_t, v_t^a, a^\pi_t, a^\sigma_t) =
\begin{cases}
    r^*_{unsafe}, & \text{if } a_t^\pi \neq a_t^\sigma \\
    r^*_{exp}, & \text{if } a_t^\pi = a_t^\sigma \ \wedge \ \rho \geq \rho^* \\
    r^{step}_t, & \text{otherwise}
\end{cases}
\end{equation}

where $r^*_{unsafe}$ is applied to discourage unsafe behaviors, $r^*_{exp}$ is a reward given when the agent has explored a sufficiently high portion of the exploration arena, and $r^{step}_t$ is a step-wise reward that measures the benefit of executing action $a_t^\phi$ for exploration. Based on Eq. \ref{eq:basic_reward}, we conduct an ablation study by analyzing eight different reward functions to evaluate the contribution and effectiveness of each term in the proposed reward SGA:

\begin{itemize}
    \item \textbf{Safety-Gated Exploration (SGE)}. This formulation penalizes unsafe actions while providing only a sparse terminal reward. Consequently, the step-wise term is set to $r^{step}_t = 0$ in Eq.~\ref{eq:basic_reward}, such that exploration is positively rewarded exclusively upon reaching the threshold $\rho^*$.
    
    \item \textbf{Safety-Gated Discovery (SGD)}. To explicitly incentivize discovery, this formulation assigns $r^{step}_t = n_d$, where $n_d$ denotes the number of cells newly discovered at step $t$, thereby capturing the progressive expansion of the known map.
    
    \item \textbf{Safety-Gated Pure Exploration (SGPE).} This approach differs from the previous ones by positively rewarding discoveries while penalizing a lack of progress in the exploration process, with the step reward $r^{step}_t$ defined in Eq. \eqref{eq:reward3_exp}.
    
    \begin{equation} 
    \label{eq:reward3_exp}
    r^{\text{step}}_t =
    \begin{cases}
        r_0, & \text{if } n_d \leq 0 \\[6pt]
        n_d, & \text{if } n_d > 0
    \end{cases}
    \end{equation}
    
    \item \textbf{Safety-Gated Adaptive (SGA)}. This reward function is the one proposed in this paper and integrates penalization of unsafe actions, discovery-driven rewards, and a potential-field-inspired term defined on frontier points. Therefore, the step-wise reward is defined according to Eq. \eqref{eq:reward4}.

    \begin{equation}
    \label{eq:reward4}
    r^{\text{step}}_t =
    \begin{cases}
        n_d, & \text{if } n_d > 0 \\
        \Phi_{t-1} - \Phi_t, & \text{if } n_d \leq 0 \ \wedge \ |\mathcal{V}^f_{t-1}|, |\mathcal{V}^f_t| > 1 \\
        r_0, & \text{otherwise} 
    \end{cases}
    \end{equation}
    
    where $\Phi_t(\mathcal{M}^a_t, v_t^a)$ is the potential field:
    
    \begin{equation} \label{eq:potential_field}
    \Phi_t = \max_{f \in \mathcal{V}^f_t} \frac{\Gamma_{unknown}(f)}{\| f - v_t^a \|_2},
    \end{equation}
    with $v_t^a$ the agent’s position, $\mathcal{V}^f_t$ the set of frontiers and $\| \cdot \|_2$ is the Euclidean distance. 

\end{itemize}

For each formulation, we additionally consider variants in which the unsafe penalty $r^*_{unsafe}$ is removed. These are denoted respectively as Free Exploration (FE), Free Discovery (FD), Free Pure Exploration (FPE), and Free Adaptive (FA). As ablation study, the eight policies based on these reward formulations have been trained and evaluated over 100 simulations in randomly generated environments, comparing map coverage and safety shield interventions to assess the impact of each reward terms.  

\section{Simulation and experiment}

\begin{table}[]
    \centering
    \caption{Parameters used for environment, training setup, and reward terms}
    \label{tbl:parameters}
    \setlength{\tabcolsep}{5pt} 
    \renewcommand{\arraystretch}{1.1} 
    \begin{tabular}{|l c|l c|l c|}
        \hline
        \multicolumn{2}{|c|}{Environment} & 
        \multicolumn{2}{c|}{Training} & 
        \multicolumn{2}{c|}{Reward} \\
        \hline
        Name & Value & 
        Name & Value & 
        Name & Value \\
        \hline
        $h, w$ & 50 & Rollouts & 1024 & $r^*_{\text{unsafe}}$ & $-1$ \\
        $n_o$ & 45 & Mini-Batches & 64 & $r_0$ & $-0.5$ \\
        $r_{\min}, r_{\max}$ & $[1,3]m$ & Learning Rate & $3\!\times\!10^{-4}$ & $r^*_{\text{exp}}$ & $100$ \\
        $l$ & 5.0 & Learning Epochs & 8 & & \\
        $\rho^*$ & 0.98 & Discount factor & 0.99 & & \\
        $n_s^*$ & 2500 & Timesteps & $4 \cdot 10^5$ & & \\
        $k$ & 3 & $\epsilon$ & 0.2 & & \\
        $r_g$ & 1.0 m& & & & \\
        \hline
    \end{tabular}
\end{table}

This section outlines the training setup in \ref{sbs:training}, and presents the evaluation of the trained policies both in Gymnasium-based simulations in \ref{sbs:results} and experiments in a lab environment \ref{sbs:experiment_results}.

\begin{figure*}[]
    \centering
    \includegraphics[width=\textwidth]{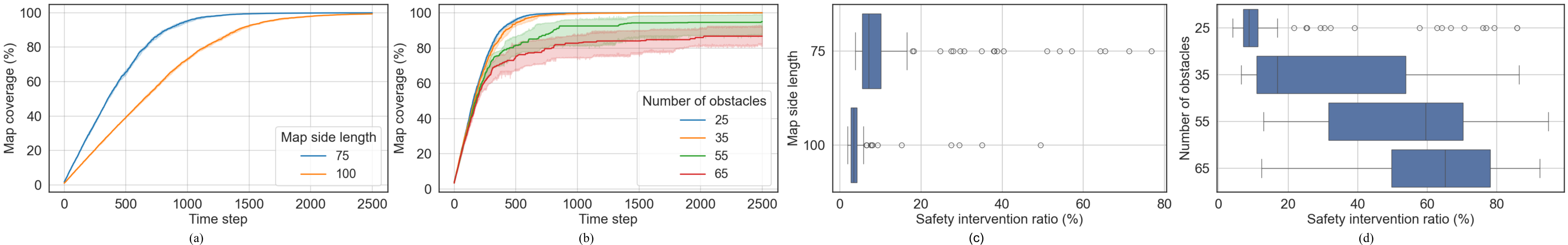}
    \caption{From left to right: median map coverage trajectories across 100 randomly generated exploration environments with different environment sizes (a) and number of trees (b) for the proposed policy SGA. Distribution of the proportion of safety filter interventions over agent actions across 100 test simulations for the policy SGA in environments with different sizes (c) and number of trees (d).}
    \label{fig:rob}
\end{figure*}

\subsection{Training setup}
\label{sbs:training}

The training is conducted on HPC2N's Kebnekaise cluster running Ubuntu 20.04.6 LTS, utilizing 5 Skylake CPU cores and an NVIDIA H100 GPU, while policy evaluations are performed on the same system using 10 Skylake CPU cores. The environment is implemented using Gymnasium \cite{towers2024gymnasium}, with the main design parameters summarized in Table \ref{tbl:parameters}. Both the exploration policy and the critic network are modeled as graph-based neural networks using PyTorch Geometric \cite{Fey/Lenssen/2019} and trained in environments with randomized initial agent placements using the PPO algorithm implemented by skrl \cite{serrano2023skrl}, with hyperparameters listed in Table \ref{tbl:parameters}. 
Fig. \ref{fig:mean_total_reward} depicts the evolution of the mean total reward during training of the policy trained using SGA, characterized by an overall upward trend with an initial phase of instability up to approximately $2\times 10^{4}$ steps, after which the curve stabilizes and converges. The policy was trained for slightly more than $5\times 10^{5}$ steps, wherein the policy converged after about $1\times 10^{5}$ steps.

\begin{figure*}
    \centering
    \includegraphics[width=\textwidth]{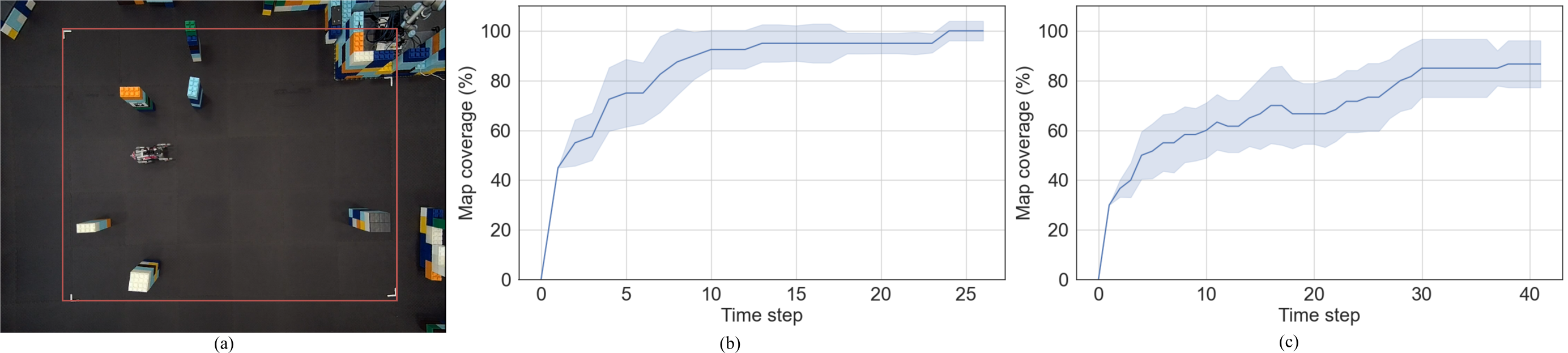}
    \caption{From left to right: (a) Custom-designed exploration arena, delineated in red, constructed in a laboratory environment, with blocks defining non-traversable regions and the Unitree Go1 quadruped robot acting as the exploring agent. Median map coverage obtained by the Unitree Go1 across 10 environments with randomly placed obstacles in arenas of size \(4 \,\text{m} \times 5 \,\text{m}\) (b) and 10 arenas of size \(6 \,\text{m} \times 5 \,\text{m}\) (c).
}
    \label{fig:exp}
\end{figure*}

\subsection{Simulation results}
\label{sbs:results}

The hierarchical exploration framework with the exploration policies trained under the devised eight reward formulations discussed in ~\ref{sbs:reward_abl} have been evaluated across 100 randomly-generated testing environments in Gymnasium, each executed for 2,500 time steps, in order to assess the robustness to variations in non-traversable region placement and agent initialization. Fig. \ref{fig:sim} shows the simulation data, including the coverage of the map as a function of the exploration time and the frequency of safety-filter interventions to prevent unsafe actions. These results provide the basis for the ablation analysis of the proposed SGA reward function.

Fig. \ref{fig:sim}-(a) illustrates the median and standard deviation of map coverage as a function of time steps during exploration for each evaluated policy. The use of the median trajectory provides a robust indicator of typical performance across environments, mitigating the influence of outliers. The policy trained with the proposed SGA reward exhibits the steepest and most consistent increase in coverage, achieving 95\% map exploration within approximately 605 steps in median performance. These results underscore the efficiency and robustness of the proposed framework, even under varying environment configurations. In contrast, policies trained with reward functions comprising fewer components than SGA yield exploration trajectories with a smaller slope, such as FPE, and, in some cases, substantially reduced map coverage considering corresponding steps. For example, after 1,000 steps, the SGA policy achieves over 99\% coverage, whereas FE remains at only 6.7\%. In particular, SGE and FE, the simplest reward formulations, yield the lowest performance, failing to exceed 60\% coverage within the first 1,000 steps. Furthermore, SGE exhibits substantial variability across simulations, as indicated by the large standard deviation reported in Fig. \ref{fig:sim}-(a). Incorporating additional reward components, as in SGD and FD, yields more favorable trajectories by explicitly encouraging incremental exploration. FD attains high levels of coverage; however, its trajectory is flatter than those of other policies and converges to lower final coverage. SGPE further improves performance by penalizing non-expansive actions. Nevertheless, SGA consistently outperforms all alternatives, achieving both faster convergence and higher overall coverage. Fig. \ref{fig:sim}-(b) presents the distribution of map coverage after 1,000 steps. The SGA policy achieves the highest coverage and exhibits the most reliable performance, as indicated by its narrow interquartile range (IQR). Although FE displays an even narrower IQR, around 2.5, its overall coverage remains extremely limited. All other policies yield inferior results, characterized by both lower median coverage and greater variability across simulations. In particular, SGE shows highly inconsistent behavior, with an IQR of 70. Fig. \ref{fig:sim}(c) shows the proportion of agent steps in which the safety filter intervenes to prevent collisions. Policies trained with reward functions that explicitly penalize unsafe actions (\textit{e.g.}, SGPE, SGE, SGD, and SGA) generally exhibit fewer interventions, as reflected by lower median filter activations, compared to formulations without such penalization. This indicates that these policies not only achieve efficient exploration but also learn to operate with reduced reliance on the safety filter. Nonetheless, the occurrence of occasional activations highlights the filter’s continued relevance, providing robust safety guarantees in critical scenarios.

To further assess robustness, the proposed SGA policy was evaluated in environments with varying numbers of obstacles and different map sizes. For each configuration, 100 simulations of 2,500 steps were conducted. Fig. \ref{fig:rob} illustrates the policy’s adaptability to both conditions. In particular, Fig. \ref{fig:rob}-(a) shows that the agent successfully explores more than 90\% of the traversable area within 831 steps in maps 125\% larger than the nominal size, and within 1,405 steps in maps 300\% larger. As shown in Fig. \ref{fig:rob}-(b), increasing the proportion of non-traversable regions slightly reduces coverage at a fixed number of steps; nevertheless, the policy maintains over 80\% coverage even with a 45\% increase in obstacles. The impact of obstacle density is further reflected in Fig. \ref{fig:rob}-(d), where safety-filter activations grow with increasing obstacle counts, underscoring the importance of the filter in ensuring safe exploration. Conversely, Fig. \ref{fig:rob}-(c) shows that the frequency of safety interventions decreases with larger environment sizes. Eventually, all results confirm the scalability and robustness of the proposed framework across diverse exploration scenarios.

\subsection{Experimental results}
\label{sbs:experiment_results}

To evaluate the trained policy in real environment, the proposed framework was implemented in ROS~2 (Humble Hawksbill) and deployed on a Unitree Go1 quadruped robot. The experiments have been carried out in multiple custom-designed exploration arenas with varying sizes, obstacle configurations, and obstacle densities, all constructed within a laboratory environment, as illustrated in Fig. \ref{fig:exp}-(a). The figure shows one configuration of the arena, outlined in red, with dimensions of \(6 \,\text{m} \times 5 \,\text{m}\). The Unitree Go1 quadruped robot is shown alongside several randomly positioned LEGO blocks, which define non-traversable regions within the arena. The Unitree Go1 robot is equipped with an onboard Intel NUC computer and a SLAMTEC RPLIDAR S1 360° Laser Scanner. Ground-truth position and orientation are obtained using a Vicon motion capture system. The implemented framework assigns to the agent a ROS~2 node that is responsible for: (i) collecting laser scans from the LiDAR and odometry measurements from the Vicon system when the robot reaches an exploration waypoint; (ii) computing the exploration graph $o_t$; (iii) inferring the action $a^\pi_t$ by applying the trained policy $\pi_\theta$ on the exploration graph $o_t$; (iv) applying the safety filter that accounts for the reconstructed map and nearby non-traversable regions, thus constraining the policy action $a_t^\pi$ to yield a feasible action $a_t^\sigma$; and (v) generating control commands to drive the quadruped toward the next waypoint. For these experiments, the LiDAR range was restricted to $2.12 \,\text{m}$, so that at each time step the robot could only assess the traversability of its neighboring cells. Moreover, each action selected by the proposed hierarchical exploration framework is executed by a navigation controller, which guides the robot to the center of the neighboring cell corresponding to the chosen action. In the case of diagonal movements, the robot sequentially traverses two adjacent cells to reach the final goal, with the constraint that such movement is permitted only if at least one of the two intermediate cells is traversable.
Fig. \ref{fig:exp}-(b) reports the map coverage across 10 experiments conducted in arenas of size \(4 \,\text{m} \times 5 \,\text{m}\) with randomly placed obstacles. The results indicate that after 7 time steps the agent achieved more than 80\% coverage, and after 13 time steps it exceeded 95\%. The framework is further evaluated in larger arenas, as illustrated in Fig. \ref{fig:exp}-(c). Specifically, 10 experiments are conducted in the larger arena of size \(6 \,\text{m} \times 5 \,\text{m}\). In this setting, the map coverage reached 60\% after 10 steps and 80\% after 28 steps.
Overall, the experimental results demonstrate that the proposed framework successfully generalizes to environments that differ from those encountered during training, thereby confirming the robustness of the SGA policy.

\section{CONCLUSIONS}

This paper investigates a novel platform-agnostic hierarchical framework for safe reinforcement learning-based exploration in obstacle-rich cluttered environments. The method combines an exploration-oriented policy implemented as a graph neural network policy, with attention-based message passing, and a safety filter that guarantees the feasibility of the chosen exploration actions. Furthermore, this article proposes a novel graph representation of the environment with exploration-relevant locations and a frontier-inspired reward design that promote movement towards high-gain frontiers. Eventually, simulation studies across 100 randomly-generated testing environments in Gymnasium and experiments in a lab environment demonstrate the devised agent's efficacy to achieve robust map coverage across diverse environments, with consistent performance under varying obstacle densities and scales. Future work will focus on deploying the framework as a high-level exploration planner in a real-world experiment.



\bibliographystyle{IEEEtran}
\bibliography{main}

\end{document}